\documentclass[sigconf, nonacm]{acmart}

\usepackage{enumitem}
\usepackage{multirow}

\newcommand\vldbyear{2026}
\newcommand\vldbworkshop{Tabular Data Analysis (TaDA)}
\newcommand\vldbauthors{\authors}
\newcommand\vldbtitle{\shorttitle} 
\newcommand\vldbavailabilityurl{\url{https://github.com/brains-group/table-agent-toolkit}}
\newcommand\vldbpagestyle{plain} 

\begin{document}
\title{PFN-Syn: Generating Synthetic Tabular Data with Prior-Data Fitted Networks}
\settopmatter{authorsperrow=4}
\author{Michael Zuo}
\affiliation{%
  \institution{RPI}
  \city{Troy}
  \state{NY, USA}
}
\email{zuom@rpi.edu}

\author{Inwon Kang}
\affiliation{%
  \institution{RPI}
  \city{Troy}
  \state{NY, USA}
}
\email{kangi@rpi.edu}

\author{Oshani Seneviratne}
\affiliation{%
  \institution{RPI}
  \city{Troy}
  \state{NY, USA}
}
\email{senevo@rpi.edu}

\author{Stacy Patterson}
\affiliation{%
  \institution{RPI}
  \city{Troy}
  \state{NY, USA}
}
\email{sep@cs.rpi.edu}

\begin{abstract}
Synthetic tabular data can help enable data sharing and model development in privacy-sensitive domains, but existing generators often require dataset-specific training and substantial computation.
We introduce PFN-Syn, a training-free framework for synthetic tabular data generation based on prior-data fitted networks.
We evaluate PFN-Syn against training-based generators on eight binary classification datasets.
PFN-Syn achieves the highest average downstream utility among evaluated synthetic generators, with higher generation throughput than training-based baselines, while maintaining competitive distributional similarity and distance-to-closest-record privacy behavior.
Our results suggest that tabular foundation models can serve as useful building blocks for practical synthetic tabular data generation.
\end{abstract}

\maketitle

\pagestyle{\vldbpagestyle}
\begingroup\small\noindent\raggedright\textbf{VLDB Workshop Reference Format:}\\
\vldbauthors. \vldbtitle. VLDB \vldbyear\ Workshop: \vldbworkshop.\\ 
\endgroup
\begingroup
\renewcommand\thefootnote{}\footnote{\noindent
This work is licensed under the Creative Commons BY-NC-ND 4.0 International License. Visit \url{https://creativecommons.org/licenses/by-nc-nd/4.0/} to view a copy of this license. For any use beyond those covered by this license, obtain permission by emailing \href{mailto:info@vldb.org}{info@vldb.org}. Copyright is held by the owner/author(s). Publication rights licensed to the VLDB Endowment. \\
\raggedright Proceedings of the VLDB Endowment. 
ISSN 2150-8097. \\
}\addtocounter{footnote}{-1}\endgroup

\ifdefempty{\vldbavailabilityurl}{}{
\vspace{.3cm}
\begingroup\small\noindent\raggedright\textbf{VLDB Workshop Artifact Availability:}\\
The source code, data, and/or other artifacts have been made available at \url{https://github.com/brains-group/table-agent-toolkit}.
\endgroup
}

\section{Introduction}
Tabular data is central to many modern AI applications in domains such as finance and healthcare, where predictive models rely on large volumes of structured data to identify patterns, support decision-making, and improve outcomes.
However, these same domains are governed by strict regulations on data privacy and sharing, such as the General Data Protection Regulation (GDPR), Health Insurance Portability and Accountability Act (HIPAA), and Fair Credit Reporting Act (FCRA).
Synthetic data generation offers a promising solution to this challenge by creating artificial datasets that preserve the statistical structure and relationships of the original data without directly exposing sensitive records~\cite{dankar2022}.

A variety of synthetic data generation methods have been developed for tabular data.
Early approaches relied primarily on statistical methods such as the Gaussian Copula~\cite{patki7796926}.
More modern techniques include generative adversarial networks (GANs)~\cite{xu2019}, variational autoencoders (VAEs)~\cite{xu2019}, diffusion-based methods~\cite{shi2025tabdiff}, and LLM-based generators~\cite{borisov2023language}.
While these methods have demonstrated strong performance in many settings, they often require significant computational resources, careful hyperparameter tuning, and repeated retraining when datasets or tasks change.

To address this challenge, we propose a \emph{training-free} alternative path to high-quality synthetic tabular data generation using a prior-data fitted network (PFN)~\cite{2022transformers}.
A PFN is a model that is pre-trained on synthetic data sampled from a prior family of distributions.
At inference time, the model takes both training data and query samples for a given task as input and performs prediction based directly on the in-context training samples.
There is no need to retrain the model for each new dataset and/or task.

In this work, we introduce PFN-Syn, a principled framework for converting a PFN-based predictor into a synthetic data generator for tabular data.
PFN-Syn inherits the benefits of PFN predictors: it is training-free and computationally efficient.
Evaluated on standard benchmark datasets, PFN-Syn achieves higher average downstream task utility than state-of-the-art training-based tabular data generators, while matching these generators on various quality metrics.

\paragraph{Related Work}

PFN-Syn builds on existing work on PFN predictors.
Recently, specialized PFNs have been developed for in-context learning and prediction on tabular data, TabPFN~\cite{tabpfn,grinsztajn2026tabpfn3technicalreport} and TabICL~\cite{qu2025tabicl,qu2026tabiclv2} being the most prominent examples.
These PFN-based predictors outperform state-of-the-art ML methods on various tabular dataset benchmarks~\cite{erickson2026tabarena}.

We apply an autoregressive approach to synthetic data generation which is conceptually similar to an unsupervised imputation-based framing described as an extension to TabPFN~\cite{tugnoli2026improvingtabpfnssyntheticdata,grinsztajn2026tabpfn3technicalreport}.
However, rather than extending imputation, our method explicitly generates synthetic records by sampling conditional distributions.

While a prior work uses a PFN backing an energy-based model to predict labels for synthetically generated samples~\cite{ma2024tabpfgentabulardata}, our method generates a fully synthetic table from a real table by using the PFN to predict all columns, conditioned on previous column values.

\paragraph{Outline}
In Section~\ref{method.sec}, we present the details of PFN-Syn.
Section~\ref{results.sec}  gives our experimental results, which explore variations on PFN-Syn and compare PFN-Syn's performance  against various benchmark generators.
Finally, we conclude in Section~\ref{conclusion.sec}.

\section{PFN-Syn} \label{method.sec}

\begin{figure*}[t]
\vspace{-1.1cm}
\includegraphics[width=\linewidth]{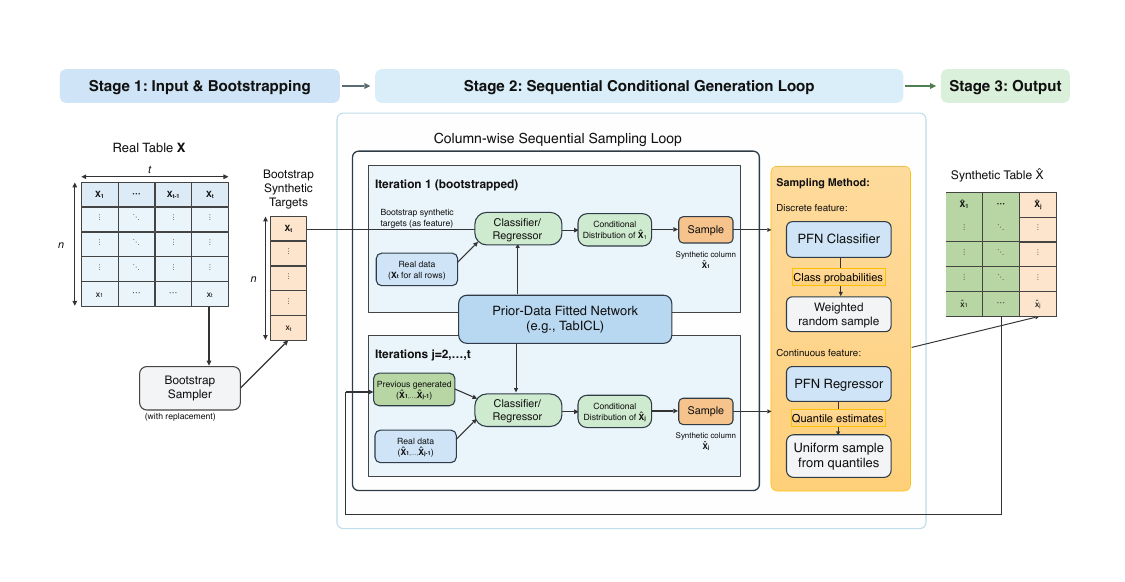}
\vspace{-1.1cm}
\caption{PFN-Syn synthetic data generation framework.} 
\label{method.fig}
\end{figure*}

Given a real tabular dataset $\mathbf{X}$ with $n$ samples and $t$ features per sample, let $\mathbf{X}_1,
\mathbf{X}_2, \ldots, \mathbf{X}_t$ be the columns of $\mathbf{X}$. We take $\mathbf{X}_t$ to be the target column; we assume target $\mathbf{X}_t$ takes on discrete values.
We seek to produce a synthetic table with the same schema and joint distribution as $\mathbf{X}$.

With a true joint distribution over all features, we could generate synthetic data simply by repeatedly sampling new rows from the distribution.
However, directly modeling full joint distributions 
is intractable for practical tabular datasets.
PFN-Syn factors the full joint distribution into a sequence of conditional sampling problems:
$$\operatorname{Pr}(\mathbf{X}_1, \ldots, \mathbf{X}_t)=\prod_{j=1}^t\operatorname{Pr}(\mathbf{X}_j~|~\mathbf{X}_1,\ldots, \mathbf{X}_{j-1}).$$
This allows us to generate one column at a time instead of modeling all columns at once.
We can now approximate each subproblem by using a PFN to estimate conditional distributions for each column given generated values in all previous columns.

To generate a synthetic row, we first bootstrap the conditional generation process by generating synthetic values $\mathbf{\hat X}_0$ according to the unconditional distribution of the target column $\mathbf{X}_t$ in the real data.
In the case where the target is a discrete categorical feature, this is equivalent to sampling directly from target column $\mathbf{X}_t$ of the real data, with replacement.
The bootstrap targets $\mathbf{\hat X}_0$ are used only when generating the first synthetic column and are then discarded; they are not included in the final output.

To generate the first synthetic column $\mathbf{\hat X}_1$, we construct a PFN training input using the real target values $\mathbf{X}_t$ as the single feature and the real first column $\mathbf{X}_1$ values as the target.
We then use the PFN to generate synthetic values for $\mathbf{\hat X}_1$ from the bootstrap targets $\mathbf{\hat X}_0$ (as a feature) by the sampling procedure described below.
For each subsequent column $j=2,\ldots,t$, we invoke the PFN using real columns $\mathbf{X}_1,\ldots,\mathbf{X}_j$ and synthetic columns $\mathbf{\hat X}_1,\ldots,\mathbf{\hat X}_{j-1}$ as input.
(The bootstrap target $\mathbf{\hat X}_0$ is not included in the query for columns $j\ge2$.)
The underlying PFN takes as input a combined table including both real and synthetic data for the fully-known columns $1,\ldots,j-1$, and the partial list of known target values $\mathbf{X}_j$.
From these inputs it generates predictions to complete the missing target values, which we use to sample the new synthetic values $\mathbf{\hat X}_j$.

The sampling procedure used to generate each column depends on whether that column is discrete or continuous.
To generate a discrete feature $\mathbf{\hat X}_j$ for each entry, we use a classifier PFN to produce a set of class probabilities conditioned on real data $\mathbf{X}_1,\ldots,\mathbf{X}_j$ and previously generated columns $\mathbf{\hat X}_1,\ldots,\mathbf{\hat X}_{j-1}$.
We determine each entry's class in the new column $\mathbf{\hat X}_j$ by sampling randomly with weights given by the classifier's reported probabilities.
The selected class is not necessarily the class predicted with the highest probability or the best-represented class in the real data, but rather is randomized to approximate the conditional distribution implied by the classifier prediction.

For continuous features, we instead use a regressor PFN fitted on $\mathbf{X}_1,\ldots,\mathbf{X}_j$ and $\mathbf{\hat X}_1,\ldots,\mathbf{\hat X}_{j-1}$ to estimate values at a set of quantiles evenly spaced within the full range $[0,1)$.
We sample values for $\mathbf{\hat X}_j$ uniformly from among these quantiles.
This procedure approximates sampling the value at a uniformly random quantile from the regressor prediction's implied distribution, drawing from the continuous feature space rather than copying any specific value from the real data.

For computational efficiency, synthetic rows are not generated individually.
We batch full sets of synthetic data so that each invocation of the underlying PFN produces the statistics required to sample all new values for a single column.
This allows for parallelism and reuse of the ``training'' corpus at each step, which is well-supported by the underlying PFN models.

\section{Evaluation Results} \label{results.sec}

\begin{table*}
\caption{Comparison of balanced accuracy of prediction from original training data and synthetic data from various generators. XGB columns show results for predictions with XGBoost, and ICL columns show results for predictions made by TabICL.} \label{accuracy.tab}
\small
\begin{tabular}{|l|cc|cc|cc|cc|cc|cc|}
\hline
\multirow{2}{*}{\bf Dataset} & \multicolumn{2}{c|}{Original} & \multicolumn{2}{c|}{Copula} & \multicolumn{2}{c|}{TabDiff} & \multicolumn{2}{c|}{CTGAN} & \multicolumn{2}{c|}{TVAE} & \multicolumn{2}{c|}{PFN-Syn} \\
    \cline{2-13}
    & XGB & ICL & XGB & ICL & XGB & ICL & XGB & ICL & XGB & ICL & XGB & ICL \\
    \hline
Adult Income (AD)~\cite{adult_2} & 0.818 & 0.844 & 0.516 & 0.507 & 0.684 & 0.640 & 0.718 & 0.711 & \textbf{0.769} & \textbf{0.758} & 0.762 & 0.684 \\\hline
Bank Customer Churn (BC)~\cite{bank_customer_churn} & 0.721 & 0.726 & 0.540 & 0.508 & 0.707 & 0.719 & 0.670 & 0.654 & 0.650 & 0.633 & \textbf{0.714} & \textbf{0.740} \\\hline
Bank Marketing (BM)~\cite{bank_marketing_222} & 0.596 & 0.594 & 0.506 & 0.5 & 0.576 & 0.580 & 0.586 & 0.589 & \textbf{0.603} & \textbf{0.604} & 0.584 & 0.578 \\\hline
Default of Credit Card Clients (CC)~\cite{default_of_credit_card_clients_350} & 0.658 & 0.659 & 0.546 & 0.503 & 0.642 & 0.648 & 0.609 & 0.623 & 0.635 & 0.631 & \textbf{0.655} & \textbf{0.653} \\\hline
German Credit Data (CR)~\cite{german_credit_data_144} & 0.686 & 0.673 & 0.533 & 0.497 & 0.614 & \textbf{0.659} & 0.507 & 0.5 & 0.552 & 0.570 & \textbf{0.616} & 0.620 \\\hline
Give Me Some Credit (GM)~\cite{GiveMeSomeCredit} & 0.594 & 0.581 & 0.502 & 0.5 & 0.583 & 0.565 & 0.628 & 0.618 & \textbf{0.666} & \textbf{0.662} & 0.640 & 0.650 \\\hline
Hazelnut Spread Contaminant Detection (HS)~\cite{ricci2021machine} & 0.897 & 0.965 & 0.803 & 0.85 & 0.872 & 0.910 & 0.566 & 0.487 & 0.785 & 0.788 & \textbf{0.873} & \textbf{0.922} \\\hline
Polish Companies Bankruptcy (PB)~\cite{polish_companies_bankruptcy_365} & 0.760 & 0.755 & 0.499 & 0.5 & 0.601 & 0.523 & \textbf{0.677} & \textbf{0.635} & 0.502 & 0.517 & 0.581 & 0.574 \\\hline
{\bf Mean} & 0.716 & 0.725 & 0.556 & 0.546 & 0.660 & 0.656 & 0.621 & 0.602 & 0.645 & 0.645 & 0.678 & 0.678 \\\hline
\end{tabular}
\end{table*}

\begin{table}
\caption{Average utility, quality, and privacy metrics over eight datasets (higher is better).}
\label{summary.tab}
\small
\begin{tabular}{|l|c|c|c|c|c|c|}
\hline
Generator & XGB & ICL & Shape & Pair & DCR-B & DCR-O \\
\hline
Original & 0.716 & 0.725 & (1) & (1) & (0) & (0) \\
\hline
Copula & 0.556 & 0.546 & 0.875 & 0.871 & 0.258 & 0.815 \\
\hline
TabDiff & 0.660 & 0.656 & \textbf{0.972} & \textbf{0.962} & 0.135 & 0.823 \\
\hline
CTGAN & 0.621 & 0.602 & 0.789 & 0.815 & \textbf{0.263} & \textbf{0.866} \\
\hline
TVAE & 0.645 & 0.645 & 0.857 & 0.802 & 0.091 & 0.786 \\
\hline
PFN-Syn & \textbf{0.678} & 
\textbf{0.678} & 0.856 & 0.841 & 0.131 & 0.855 \\
\hline
\end{tabular}
\end{table}

\begin{figure}[t]
\caption{Comparison of data generators by throughput, plotted log-log against training set size (higher is better).} 
\label{throughput.fig}
\includegraphics[width=\linewidth]{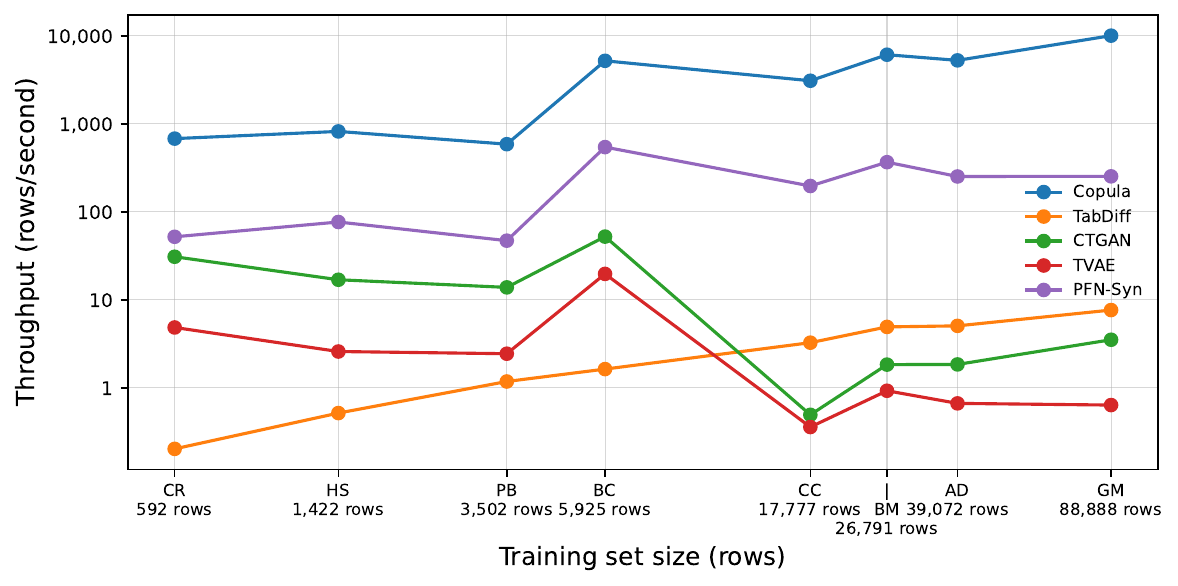}
\end{figure}

\paragraph{Setup}
We instantiate PFN-Syn using TabICL v2 to provide the underlying PFN.
We compare PFN-Syn against Gaussian Copula, TabDiff, CTGAN, and TVAE in their reference configurations on eight binary classification tabular datasets (see Table~\ref{accuracy.tab}).
Each generator produces a synthetic table of the same size as the original table.
We perform experiments on an Nvidia RTX A6000 GPU with 48 GB memory, under the conditions given in \cite{Zuo2026}, averaging results over~3 generated datasets in each configuration.

\subsection{Quality and Utility}
We measure the utility of the resulting synthetic datasets by training downstream XGBoost and TabICL classifiers on the synthetic data produced by each generator and evaluating balanced accuracy on held-out real data.
These results are shown in Table~\ref{accuracy.tab}.

PFN-Syn, while not uniformly dominant, is consistently competitive across all datasets. 
All other methods except TabDiff suffer a discriminative collapse on some dataset, achieving balanced accuracy comparable to random guessing.

Table~\ref{summary.tab} reports average scores across all datasets for balanced accuracy with the XGBoost and ICL classifiers, as well as column shapes and column pair trends data quality metrics and the distance to closest record baseline (DCR-B) and overfitting (DCR-O) protection metrics from SDMetrics~\cite{sdmetrics}.
PFN-Syn achieves the highest average downstream balanced accuracy among the synthetic generators under both downstream classifiers.

On quality and protection metrics, PFN-Syn is solidly middle-of-the-pack.
On the column shape and pair trends metrics, PFN-Syn scores within the range of the training-based generators, suggesting that it preserves marginal feature distributions comparably well.
The DCR protection metrics suggest that PFN-Syn's strong performance on balanced accuracy is not attributable to merely reproducing the real data. PFN-Syn does not generate synthetic records that are unusually similar to training records compared to the other generators.

\begin{table}
\caption{Comparison of XGBoost balanced accuracy between baseline and scalable variants.}
\label{scaling.tab}
\small
\begin{tabular}{|l|c|c|c|c|c|}
\hline
\multirow{2}{*}{\bf Dataset} & \multirow{2}{*}{Full size} & \multicolumn{2}{c|}{Split} & \multicolumn{2}{c|}{Coreset} \\
    \cline{3-6}
    & & 3:1 & 10:1 & 3:1 & 10:1 \\
    \hline
AD & 0.762 & 0.744 & 0.766 & 0.768 & 0.667 \\\hline
BC & 0.714 & 0.723 & 0.685 & 0.712 & 0.736 \\\hline
BM & 0.584 & 0.592 & 0.593 & 0.585 & 0.592 \\\hline
CC & 0.655 & 0.665 & 0.655 & 0.662 & 0.662 \\\hline
CR & 0.616 & 0.579 & 0.586 & 0.571 & 0.543 \\\hline
GM & 0.640 & 0.628 & 0.624 & 0.632 & 0.641 \\\hline
HS & 0.873 & 0.860 & 0.862 & 0.852 & 0.833 \\\hline
PB & 0.581 & 0.591 & 0.572 & 0.621 & 0.597 \\\hline
{\bf Mean} & 0.678 & 0.673 & 0.668 & 0.675 & 0.659 \\\hline
\end{tabular}
\end{table}

\subsection{Scalability}

PFN-Syn's strong performance is particularly appealing as it is very fast compared to training-based generators.
Figure~\ref{throughput.fig} plots estimated synthetic data generation throughput of each generator when used to generate synthetic datasets of equal size to the training data. For training-based generators, timings include training and generation time.
PFN-Syn is outperformed only by Gaussian Copula, a statistical generator with poor output utility.
Compared to generators based on dataset-specific model training, PFN-Syn achieves 1.7--74 times higher throughput while maintaining competitive data quality and downstream utility, qualities that make PFN-Syn attractive for rapid exploratory analysis and practical synthetic data generation on small to moderately large datasets.

PFN-Syn inherits the context and query size constraints of its PFN substrate; the TabICL v2 models on which our experiments are based report good performance for up to 100,000 samples, requiring disk offloading for larger datasets due to an intermediate memory bottleneck~\cite{qu2026tabiclv2}.
However, PFN-Syn also stands to benefit from improvements in the underlying foundation models; the recent TabPFN-3 claims good performance up to 1 million rows with small feature counts~\cite{grinsztajn2026tabpfn3technicalreport}.

As existing PFNs are primarily memory-constrained on total input size for a single invocation, we explore two approaches to scaling PFN-Syn by applying the underlying PFN model on smaller or partial inputs.

In the ``split'' settings, we randomly divide the samples evenly in each dataset into disjoint batches, using 3 and 10 batches in total, respectively.
We then apply PFN-Syn to generate synthetic data from each batch equal to the smaller batch size.
Concatenating the synthetic outputs from each batch results in a total output size equal to the original dataset.
While the total amount of input to the underlying PFN does not decrease, the input to each individual PFN invocation is smaller.
Moreover, as there is no interaction between the samples in each batch, data generation from each batch can be performed in parallel.

In the ``coreset'' settings, we apply K-means clustering to select a representative coreset smaller than the original dataset.
We find a number of clusters equal to 1/3 and 1/10 of the dataset size, respectively, then use the cluster centroids as an input dataset for PFN-Syn.
This approach does reduce the total amount of input to the underlying PFN, but requires a variable amount of precomputation, depending on the coreset selection method.

We present the performance impact of these variations in Table~\ref{scaling.tab}. While we observe the intuitively expected trend of reduced utility as we reduce per-input size overall, the empirical impact is small and performs inconsistently between datasets.
This limited performance degradation suggests that scaling of the underlying PFN models is not fundamentally problematic for synthetic data generation.

\begin{table}
\caption{Comparison of balanced accuracy between baseline and randomized feature generation orders.}
\label{order.tab}
\small
\begin{tabular}{|l|cc|cc|cc|}
\hline
\multirow{2}{*}{\bf Dataset} & \multicolumn{2}{c|}{In-Order} & \multicolumn{2}{c|}{Target Last} & \multicolumn{2}{c|}{Full Random} \\
    \cline{2-7}
    & XGB & ICL & XGB & ICL & XGB & ICL \\
    \hline
AD & 0.762 & 0.684 & 0.693 & 0.663 & 0.705 & 0.643 \\\hline
BC & 0.714 & 0.740 & 0.713 & 0.729 & 0.723 & 0.709 \\\hline
BM & 0.584 & 0.578 & 0.593 & 0.579 & 0.575 & 0.582 \\\hline
CC & 0.655 & 0.653 & 0.658 & 0.658 & 0.650 & 0.643 \\\hline
CR & 0.616 & 0.620 & 0.613 & 0.635 & 0.609 & 0.625 \\\hline
GM & 0.640 & 0.650 & 0.654 & 0.626 & 0.640 & 0.630 \\\hline
HS & 0.873 & 0.922 & 0.879 & 0.920 & 0.872 & 0.910 \\\hline
PB & 0.581 & 0.574 & 0.553 & 0.535 & 0.598 & 0.627 \\\hline
{\bf Mean} & 0.678 & 0.678 & 0.670 & 0.668 & 0.672 & 0.671 \\\hline
\end{tabular}
\end{table}

\subsection{Feature Generation Ordering}

We examine the impact of feature generation order on downstream performance by comparing PFN-Syn's straightforward in-order generation scheme with randomly shuffled column orders.

The in-order setting is baseline PFN-Syn, as previously described.
In the ``target last'' setting, we randomly permute the order of non-target features, but ensure that the target feature of the downstream task is used as the bootstrap target and generated as the final column.
In the ``full random'' setting, we include the target column in the permutation, allowing it to be generated at any point during the sampling process.
Each trial of the random settings uses a separate column permutation.

We compare the performance of these variations in Table~\ref{order.tab}.
While other work suggests that causal structure has an impact on synthetic data quality~\cite{tugnoli2026improvingtabpfnssyntheticdata}, we do not observe a significant difference in downstream task performance between in-order generation and randomly permuted orders,
nor do we find any clear pattern where column order randomization degrades performance.

\section{Conclusion} \label{conclusion.sec}

PFN-Syn reframes synthetic data generation as a sequence of in-context conditional prediction problems.
Instead of training a new generator for each table, we repurpose a pre-trained tabular prior-data fitted network as a conditional sampler.
Despite the simplicity of its design, PFN-Syn provides synthetic data with extremely strong downstream predictive utility on classification tasks across eight benchmark datasets, along with competitive distributional similarity and DCR-based privacy scores, while generating data very fast compared to training-based approaches.
Our results suggest that pre-trained tabular foundation models are useful not only for prediction, but can also serve as useful building blocks for an efficient, practical alternative to training-based synthetic tabular data generation.

\begin{acks}
Authors acknowledge the support from NSF IUCRC CRAFT center research grant (CRAFT Grant \#22023) for this research.
The opinions expressed in this publication do not necessarily represent the views of NSF IUCRC CRAFT.
\end{acks}

\bibliographystyle{ACM-Reference-Format}
\bibliography{references}

\end{document}